\documentclass{article}

\usepackage[accepted]{icml2023}
\usepackage{amsmath}
\usepackage{amssymb}
\usepackage{multirow}
\usepackage{booktabs}
\usepackage{tabularx}
\usepackage{bbm}
\usepackage{authblk} 
\title{On Adversarial Robustness and Out-of-Distribution Robustness of Large Language Models}

\author[]{\text{April Yang}}
\author[]{\text{Jordan Tab}}
\author[]{\text{Parth Shah}}
\author[]{\text{Paul Kotchavong}}
\affil[]{Carnegie Mellon University}

\date{}

\usepackage[disable,textsize=tiny]{todonotes}
\usepackage[textsize=tiny]{todonotes}

\icmltitlerunning{Investigating the Relationship between Adversarial and OOD Robustness in Foundation Models}

\begin{document}

\twocolumn[
\maketitle








\icmlkeywords{Large Language Models, Adversarial Robustness, Out-of-Distribution Robustness, Machine Learning}

\vskip 0.3in
]


\begin{abstract}
The increasing reliance on large language models (LLMs) for diverse applications necessitates a thorough understanding of their robustness to adversarial perturbations and out-of-distribution (OOD) inputs. In this study, we investigate the correlation between adversarial robustness and OOD robustness in LLMs, addressing a critical gap in robustness evaluation. By applying methods originally designed to improve one robustness type across both contexts, we analyze their performance on adversarial and out-of-distribution benchmark datasets. The input of the model consists of text samples, with the output prediction evaluated in terms of accuracy, precision, recall, and F1 scores in various natural language inference tasks. 

Our findings highlight nuanced interactions between adversarial robustness and OOD robustness, with results indicating limited transferability between the two robustness types. Through targeted ablations, we evaluate how these correlations evolve with different model sizes and architectures, uncovering model-specific trends: smaller models like LLaMA2-7b exhibit neutral correlations, larger models like LLaMA2-13b show negative correlations, and Mixtral demonstrates positive correlations, potentially due to domain-specific alignment. These results underscore the importance of hybrid robustness frameworks that integrate adversarial and OOD strategies tailored to specific models and domains. Further research is needed to evaluate these interactions across larger models and varied architectures, offering a pathway to more reliable and generalizable LLMs.
\end{abstract}

\section{Introduction}

Large Language Models (LLMs) are increasingly used for diverse natural language processing tasks but face significant challenges in maintaining reliability when exposed to adversarial perturbations and out-of-distribution (OOD) inputs. While methods for improving adversarial and OOD robustness have been developed independently, the relationship between these two robustness types remains underexplored. To address this gap, our study systematically evaluates the correlation between adversarial and OOD robustness across foundation models of varying architectures and parameter sizes.

We evaluate three models—Llama2-7b, Llama2-13b, and Mixtral-8x7b—on four benchmark datasets. For adversarial robustness, we utilize PromptBench and AdversarialGLUE++, and for OOD robustness, we employ Flipkart and DDXPlus. These datasets span natural language inference tasks such as sentiment analysis and question answering. Key metrics, including accuracy, precision, recall, and F1 scores, are calculated to assess performance.

By implementing and testing two robustness improvement strategies—Analytic Hierarchy Process (AHP), designed for adversarial robustness, and In-Context Rewriting (ICR), designed for OOD robustness—we analyze their effectiveness across both contexts. This allows us to explore whether improvements in one domain generalize to the other and to speculate on how architectural differences and parameter sizes influence these correlations.

\section{Literature Review}
\subsection{Adversarial Robustness}
Adversarial robustness refers to a model's resilience against intentionally crafted input perturbations designed to mislead the model. While large-scale pre-trained LLMs have achieved impressive performance across natural language understanding (NLU) tasks, research has shown that these models can be vulnerable to carefully crafted adversarial examples \cite{wang2021adversarial}, \cite{nie2020adversarial}. Evaluating and understanding this robustness is critical, and the field has seen efforts to create standardized benchmarks to track progress and identify promising ideas.  

The authors of the Adversarial NLI paper \cite{nie2020adversarial} created the ANLI dataset by having human annotators identify and exploit weaknesses in state-of-the-art models, leading to a more challenging and robust benchmark for NLU. The ANLI dataset consists of three rounds of increasingly difficult examples, and was constructed using longer contexts sourced from multi-sentence material \cite{nie2020adversarial}. The paper demonstrated that models trained on the ANLI dataset showed significant improvements in their ability to handle adversarial examples compared to models that were not specifically trained on such data. 

Similarly, AdvGLUE is a multi-task benchmark designed to evaluate the robustness of LLMs to adversarial prompts across a range of NLU tasks. It employs 14 textual adversarial attack methods and incorporates human validation to ensure the quality of adversarial examples \cite{wang2021adversarial}. Both benchmarks highlight the challenges of evaluating robustness and the need for standardized and reliable methods to assess the true robustness of language models. We will be using an enhanced version of this benchmark called AdvGLUE++ in our evaluation. PromptRobust builds on these efforts, focusing specifically on the robustness of LLM prompts across diverse tasks, recognizing that prompts can significantly impact LLM outcomes. It uses various adversarial attacks and analyzes LLM attention weights to understand the vulnerabilities associated with prompts \cite{zhu2024promptrobustevaluatingrobustnesslarge}. These benchmarks represent important steps towards a more systematic and comprehensive evaluation of adversarial robustness in language models.

\subsection{Out-of-Distribution Robustness}
OOD robustness measures a model's performance on data distributions that differ from its training distribution. This aspect of robustness is particularly crucial for real-world applications where models encounter novel or unexpected inputs. \citet{krueger2021out} proposed Risk Extrapolation (REx), a method that aims to learn stable correlations across training environments to improve OOD generalization. \citet{wang2023robustnesschatgptadversarialoutofdistribution} evaluates the OOD robustness of several LLMs, including ChatGPT, using a zero-shot approach. This involves directly applying pre-trained models to the test data without any fine-tuning or context examples. The study leverages datasets such as Flipkart and DDXPlus to assess the models' ability to generalize to unseen domains like product reviews and medical diagnosis. The evaluation in \cite{wang2023robustnesschatgptadversarialoutofdistribution} highlights the importance of OOD robustness in practical applications, especially in safety-critical domains like medical diagnosis. The authors acknowledge that while large language models show promise in handling OOD data, their absolute performance is still far from perfect, indicating a need for further research in this area.

\subsection{Robustness Enhancement Methods}
Recent research has proposed various approaches to improve model robustness. While a simplest approach might be to finetune or retrain models on adversarial or OOD data, this can be costly and resource intensive. Several promising methods have emerged, including the Analytic Hierarchy Process for enhancing adversarial robustness \cite{liu2023enhancing}, consistency alignment training for improving response reliability \cite{zhao2024improvingrobustnesslargelanguage}, and In-Context Rewriting (ICR) techniques for OOD black-box robustness \cite{obrien2024improvingblackboxrobustnessincontext}.

Analytic Hierarchy Process (AHP) framework breaks down the task of robust inference into manageable subtasks, prioritizing them, and addressing them systematically. AHP relies on feedback from another LLM, removing the need for retraining or optimization, and has been shown to improve robustness against jailbreak attacks and adversarial attacks on downstream tasks \cite{liu2023enhancing}. Evaluation reveals that larger LLMs used as the AHP engine consistently led to greater robustness in the inference LLM.

In-context Rewriting (ICR) uses an LLM to rewrite out-of-distribution (OOD) inputs to be more stylistically similar to in-distribution (ID) data. This can improve performance, as it makes the OOD input more similar to the data the LLM was trained on. For example, using ICR with a Stable Beluga 2 LLM to augment inputs for BERT, T5, and Falcon models resulted in improved accuracy on various NLP tasks \cite{obrien2024improvingblackboxrobustnessincontext}. However, analysis suggests that while ICR can improve performance on OOD data, there is room for improvements between the ID and OOD data distributions. 

Consistency Alignment Training (CAT) aims to improve the consistency of LLM outputs when presented with semantically equivalent instructions phrased differently. This is a two-stage training process. In the first stage, the model is fine-tuned on a dataset augmented with paraphrased instructions. The second stage, response consistency alignment, involves training the model to discern subtle differences in responses to similar instructions and align outputs with human expectations \cite{zhao2024improvingrobustnesslargelanguage}. Evaluation shows that consistency alignment improves both consistency and overall performance on instruction-following tasks. \\

While existing methods effectively address adversarial and OOD robustness individually, the relationship between these robustness types remains underexplored, particularly in large-scale foundation models. This project bridges this gap by applying AHP and ICR across adversarial and OOD benchmarks to evaluate their cross-context effectiveness and resulting correlation.

\section{Experiment Setup and Workflow}

Our study evaluates the robustness of three large language models—Llama2-7b, Llama2-13b, and Mixtral-8x7b—using four benchmark datasets: PromptBench and AdversarialGLUE++ for adversarial robustness, and Flipkart and DDXPlus for out-of-distribution (OOD) robustness. These benchmarks cover diverse natural language inference tasks, such as sentiment analysis and question answering.

The experiment consists of two main phases:
\begin{enumerate}
    \item \textbf{Baseline Evaluation:}
    \begin{itemize}
        \item Each model was evaluated on all four benchmarks without any robustness enhancements to establish baseline metrics.
        \item For each sample in the benchmarks, the corresponding task prompt was combined with the sample text and provided as input to the model for inference.
        \item The model outputs were used as the predictions to calculate performance. Performance was measured using accuracy, precision, recall, and F1 scores.
    \end{itemize}
    \item \textbf{Robustness Improvement Evaluation:}
    \begin{itemize}
        \item Two robustness strategies were implemented: Analytic Hierarchy Process (AHP), designed for adversarial robustness, and In-Context Rewriting (ICR), designed for OOD robustness.
        \item Each model-strategy pairing was evaluated across all benchmarks to assess cross-context robustness.
    \end{itemize}
\end{enumerate}

\begin{figure}[h!]
    \centering
    \includegraphics[width=\columnwidth]{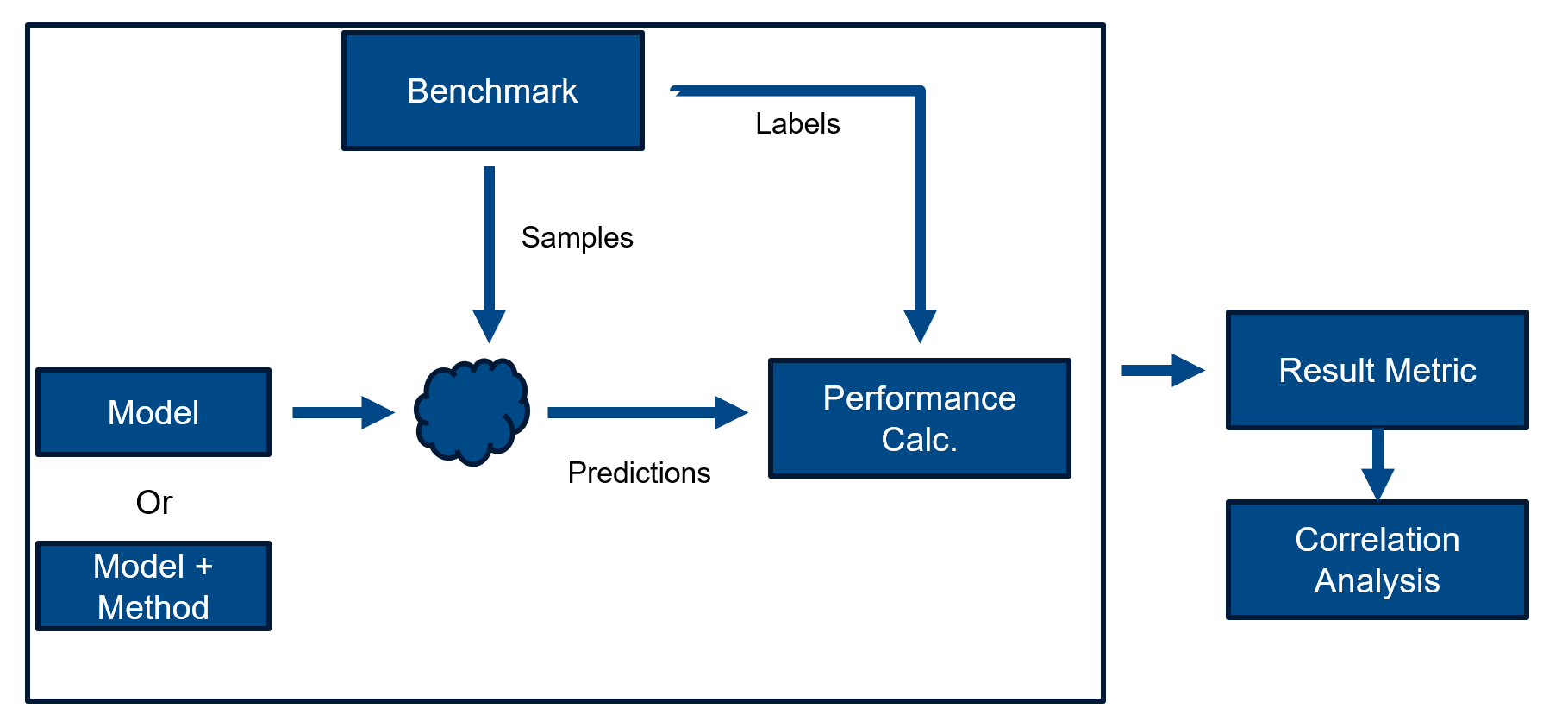}
    \caption{Entire Experiment Workflow}
    \label{fig:approach}
\end{figure}

For each model, the performance metrics from all model-strategy-benchmark pairings served as the data points for graphs combining adversarial and OOD benchmark results. Using linear regression, we derived a line of best fit to calculate correlation coefficients, providing insights into the relationship between adversarial and OOD robustness. This workflow was repeated for all models, enabling comparisons based on architecture and parameter size. Figure \ref{fig:approach} diagram shows the overall workflow. 

\subsection{Model Selection}
Our study focuses on three large language models that represent the current state-of-the-art in natural language processing. These models were chosen for their diverse architectures and training approaches, allowing us to explore how different design choices affect robustness. 

The first model is Llama-2-7b, a 7 billion parameter powerful open-source language model developed by Meta AI, known for its efficiency and performance across various NLP tasks \cite{touvron2023llama}. The second model is Llama-2-13b, an extended version of the Llama-2 series with 13 billion parameters, providing insights into the effects of scaling model size. The third model is Mixtral 8x7b, a sparse mixture-of-experts model developed by Mistral AI, offering strong performance while maintaining computational efficiency \cite{jiang2023mixtral}. The model has 45 billion parameters and a more complex architecture compared to the LLaMA-2 models. These models represent different approaches to LLM design, with Llama-2 models using a traditional transformer architecture and Mixtral employing a sparse mixture-of-experts framework. This diversity enables us to investigate how architectural choices and parameter sizes impact both adversarial and OOD robustness.

\subsection{Benchmarking Datasets}
To evaluate adversarial and OOD robustness, we selected comprehensive benchmarks designed to challenge LLMs across diverse natural language inference tasks. 

\textbf{Adversarial Robustness:}  
For adversarial robustness, we utilize two benchmarks:  
\begin{itemize}
\item \textbf{PromptRobust} \cite{zhu2024promptrobustevaluatingrobustnesslarge}: Designed to evaluate robustness to prompt variations, this benchmark systematically tests how models handle paraphrasing, noise, style changes, and varying instruction formats. Prompt perturbations include methods such as:

\begin{itemize}
    \item \textit{Semantic Attacks}: Altering sentence phrasing while preserving meaning.  
    \item \textit{Textbugger}: Character-level changes, such as misspellings or symbol replacements.
    \item \textit{Textfooler}: Replacing words with synonyms or similar words that alter the sentence's structure.
    For example:
    \begin{itemize}
        \item Original: ``Classify the emotion of this sentence as 'positive' or 'negative'.''
        \item Modified: ``Classify the sentiment of this sentence as 'positive' or 'negative'.''
    \end{itemize}
\end{itemize}

\textbf{Assumptions and Subsampling Method:}
Due to computational limitations, the SST-2 dataset was sampled to obtain a subset. The sampling maintained a proportional representation of the attack types from the overall dataset.



    \item \textbf{AdvGLUE++} \cite{wang2024decodingtrustcomprehensiveassessmenttrustworthiness}: This enhanced benchmark expands upon AdvGLUE \cite{wang2021adversarial}, incorporating more realistic adversarial examples generated through automated and human-in-the-loop methods. Tasks include sentiment analysis, question answering, and inference, with adversarial techniques such as semantic attacks, textbugger, and textfooler to challenge model robustness. An example textbugger attack:
    \begin{itemize}
        \item Original: ``Classify the emotion of this sentence as 'positive' or 'negative'.''
        \item Modified: ``Clasify th3 3motion of th1s s3nt3nc3 as 'positive' or 'negative'.''
    \end{itemize}
    \textbf{Assumptions and Subsampling Method:} Equal number of samples from each attack type and total number of samples same as PromptRobust. 
\end{itemize}


\textbf{OOD Robustness:}  
For OOD robustness, we select datasets distinct from the models' training distributions, targeting domains with significant divergence from standard NLP datasets:  
\begin{itemize}
    \item \textbf{Flipkart} \cite{flipkart}: An e-commerce dataset containing product reviews. This dataset tests real-world applications in a domain likely underrepresented in training data, emphasizing robustness in commercial contexts.  
    \\Sampling Strategy: Only considered reviews with lengths between 150-160 characters. Of the reviews with matching lengths (\textasciitilde1200 samples), considered only the first 300 for our analysis due to compute limitations and higher expected time of execution for larger datasets. This subset maintained the same distribution of labels as the overall dataset.
    \item \textbf{DDXPlus} \cite{tchango2022ddxplusnewdatasetautomatic}: A dataset of medical diagnostic conversations and clinical reasoning. This domain poses significant challenges due to its specialized language, high stakes, and divergence from general-purpose text.  
    \\Sampling Strategy: The dataset consists of 100 samples. No sampling required.

\textbf{OOD Benchmark Assumptions:} We assume that both benchmark datasets are out-of-distribution for the models we experimented with.
\end{itemize}

The combination of adversarial and OOD benchmarks ensures a comprehensive evaluation of both robustness types across our selected models.

\section{Evaluation Metrics}
We will evaluate the two types of robustness by calculating accuracy and F1-scores for each model on our benchmarking datasets. Changes in these metrics across different experiments will serve as the foundation for our robustness analysis. To quantify the relationship between adversarial and OOD robustness, we will employ linear regression. These metrics will allow us to assess the strength and direction of the relationship between different robustness measures across our selected models and datasets. 

\subsection{Metrics and Equations}

\subsubsection{Accuracy}
\begin{equation}
\text{Accuracy} = \frac{N_{\text{correct}}}{N_{\text{total}}}
\end{equation}
\textbf{Variables:}
\begin{itemize}
    \item \( N_{\text{correct}} \): Number of correctly classified examples in the zero-shot setting.
    \item \( N_{\text{total}} \): Total number of examples evaluated.
\end{itemize}

\subsubsection{F1-Score}
\begin{equation}
F1 = 2 \cdot \frac{\text{Precision} \cdot \text{Recall}}{\text{Precision} + \text{Recall}}
\end{equation}
\textbf{Variables:}
\begin{itemize}
    \item \textbf{Precision}: \( \text{Precision} = \frac{TP}{TP + FP} \)
        \begin{itemize}
            \item \( TP \): True positives (correctly predicted positive examples).
            \item \( FP \): False positives (incorrectly predicted positive examples).
        \end{itemize}
    \item \textbf{Recall}: \( \text{Recall} = \frac{TP}{TP + FN} \)
        \begin{itemize}
            \item \( FN \): False negatives (incorrectly predicted negative examples).
        \end{itemize}
\end{itemize}



\subsubsection{Linear Regression}
\begin{equation}
Y = \beta_0 + \beta_1 X + \epsilon
\end{equation}
\textbf{Variables:}
\begin{itemize}
    \item \( Y \): Dependent variable (Adversarial Metric).
    \item \( \beta_0 \): Intercept of the regression line.
    \item \( \beta_1 \): Slope of the regression line (coefficient for the independent variable).
    \item \( X \): Independent variable (OOD Metric).
    \item \( \epsilon \): Error term (residuals).
\end{itemize}
Essentially, we aim to estimate a linear relation for data points where X values are adversarial benchmark metrics and Y values are OOD benchmark metrics. 

\section{Baseline Evaluation}
\subsection{Baseline Summary}

To establish a performance baseline, we evaluated our three models—Llama2-7b, Llama2-13b, and Mixtral-8x7b—on four benchmark datasets: PromptBench and AdvGLUE++ for adversarial robustness, and Flipkart and DDXPlus for out-of-distribution (OOD) robustness. Each model was tested without any robustness enhancements to measure their default performance.

The models were used to make predictions for each sample in all benchmarks, and these predictions were compared to the ground truth labels to calculate accuracy, precision, recall, and F1 scores. The evaluation focused on natural language inference tasks, including sentiment analysis, medical diagnosis (text classification), and question answering. These baseline metrics serve as a point of comparison for assessing the impact of robustness improvement strategies implemented later in the study.

\section{Improvement Methods}
\subsection{Analytic Hierarchical Process}

The Analytic Hierarchy Process (AHP) framework approaches robustness enhancement by conceptualizing defense as a cognitive process for handling complex user queries. Drawing inspiration from hierarchical processing theory, AHP decomposes intricate tasks into manageable subtasks and systematically addresses them through a structured decision-making process. The framework consists of two main components, shown in Figure 2 below:

\begin{figure}[h]
\centering
\includegraphics[width=0.95\columnwidth]{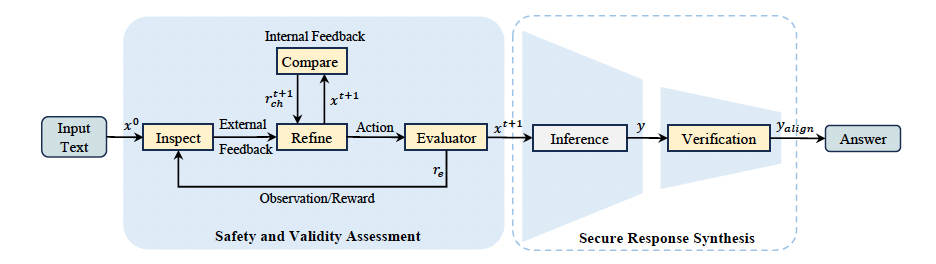}
\caption{Overview of AHP framework. The framework consists of two key components: (1) Safety and Validity Assessment, which iteratively refines input through inspection, refinement, and evaluation cycles guided by internal and external feedback; and (2) Secure Response Synthesis, which generates and verifies appropriate model outputs.}
\label{fig:ahp_framework}
\end{figure}

The Safety and Validity Assessment component operates as an iterative process to ensure input quality and safety. The system first inspects input text for potential adversarial elements (like misspellings or jailbreak templates), then refines it to remove identified issues while preserving meaning. A comparison mechanism and evaluator work together to assess the refinements and determine when the text is ready for further processing, leveraging an internal feedback loop to guide improvements.

The Secure Response Synthesis component then takes the validated input through a two-stage process to generate appropriate outputs. First, the input and the underlying task are used to generate a potential response. After the initial inference stage produces a model response, a verification step ensures this output adheres to safety requirements and task-specific formatting needs before delivering the final aligned answer. If the safety check fails, this step outputs an appropriate answer that the LLM considers "safe". This dual-check approach helps maintain output quality and safety while preserving the intended functionality.

Our implementation of the Analytic Hierarchy Process (AHP) closely followed the methodology outlined in \cite{liu2023enhancing}. For adversarial benchmarks, we utilized the same prompts from the paper. These prompts allow the LLM to iteratively refine noisy or potentially malicious adversarial tokens, reducing the impact of adversarial attacks. As for out-of-distribution benchmarks, we experimented with both the original adversarial prompts, as well as our own versions of the prompts adapted for out-of-distribution token detection and refinement. A detailed view of the exact prompts used can be found in Table \ref{tab:ahp_noise} in the appendix. For simplicity, we used the same LLM for all steps in the process.


\subsection{In-Context Rewriting} 
\textbf{In-Context Rewriting (ICR)} is a specific method within the broader framework of \textbf{LLM-powered Test-Time Augmentation (LLM-TTA)}, which uses large language models to rewrite inputs at test time before passing them to downstream tasks. This approach aims to improve robustness, particularly for out-of-distribution (OOD) inputs.

\begin{figure}[h!]
    \centering
    \includegraphics[width=\columnwidth]{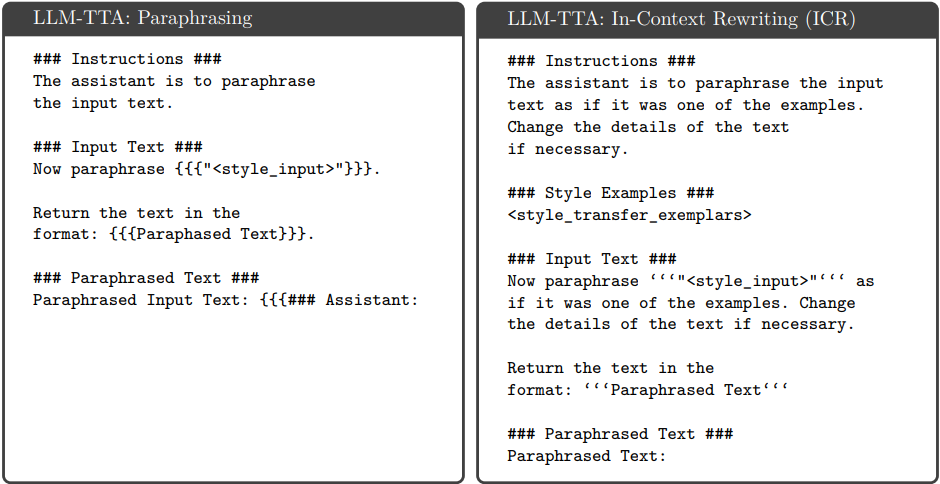}
    \caption{\textbf{LLM-TTA} Prompts. We leverage LLM-TTA with two prompting methods. The placeholder ``style\_input'' is replaced with the test input. 
    To experiment with out-of-distribution (OOD) and adversarial inputs, 
    ``style\_transfer\_exemplars'' is replaced with the ID or adversarial examples during the ICR task.}
    \label{fig:icrprompt}
\end{figure}

Rather than altering the task model itself, LLM-TTA enhances the input data itself. Given an original input \(\mathbf{x}\), LLM-TTA generates an augmented input \( \mathbf{x'} \sim \text{LLM}(\mathbf{x'} \mid P, \mathbf{x}) \)
conditioned on a natural language prompt \(P\) as well as the original input x. The two prompt templates for \(P\) is shown in Figure \ref{fig:icrprompt}. 

For our project, beyond applying In-Context Rewriting to OOD benchmarks, we also tested it on Adversarial Benchmarks. This may help the model normalize or correct adversarial inputs at test time, given examples. As shown in Figure \ref{fig:icrprompt}, there are two types of rewriting prompts, but we focus primarily on the In-Context Rewriting prompt.
\begin{itemize}
    \item \textbf{In-Context Rewriting.} In this approach, we prompt the LLM to rewrite the given OOD example to be more like a set of ID examples. For adversarial inputs, the LLM was additionally hinted to normalize and correct the perturbations in input with examples of perturbations. The LLM is expected to infer the differences between examples and \(\mathbf{x}\) via in-context learning by itself in generating augmented input \(\mathbf{x'}\).
\end{itemize}

\section{Results}
\subsection{Adversarial Benchmarks}
Tables \ref{tab:ahp_pb_sst2} and \ref{tab:ahp_plusplus} show the results for experiments evaluated on the adversarial benchmarks for all models and improvement scenarios.



\begin{table}[H]
    \centering
    \begin{tabular}{l@{\hspace{4pt}}l@{\hspace{4pt}}ccc}
        \toprule
        Model & Metric & Baseline & AHP & ICR \\ 
        \midrule
        \multirow{4}{*}{LLAMA2: 7b} 
            & Acc & .707 & .615 & .703 \\ 
            & Prec & .724 & .682 & .648 \\ 
            & Rec & .707 & .615 & .924 \\ 
            & F1 & .703 & .569 & .761 \\ 
        \midrule
        \multirow{4}{*}{LLAMA2: 13b} 
            & Acc & .774 & .583 & .809 \\ 
            & Prec & .780 & .651 & .742 \\ 
            & Rec & .774 & .583 & .953 \\ 
            & F1 & .772 & .531 & .834 \\ 
        \midrule
        \multirow{4}{*}{Mixtral: 8x7b} 
            & Acc & .802 & .855 & .692 \\ 
            & Prec & .903 & .858 & .641 \\ 
            & Rec & .649 & .855 & .909 \\ 
            & F1 & .745 & .855 & .751 \\ 
        \bottomrule
    \end{tabular}
    \caption{PromptRobust: Model Performance on SST-2 (Averaged Across Attacks)}
    \label{tab:ahp_pb_sst2}
\end{table}

\begin{table}[H]
    \centering
    \begin{tabular}{l@{\hspace{4pt}}l@{\hspace{4pt}}ccc}
        \toprule
        Model & Metric & Baseline & AHP & ICR \\ 
        \midrule
        \multirow{4}{*}{LLAMA2: 7b} 
            & Acc & .441 & .453 & .517 \\ 
            & Prec & .412 & .416 & .515 \\ 
            & Rec & .596 & .453 & .685 \\ 
            & F1 & .472 & .399 & .551 \\ 
        \midrule
        \multirow{4}{*}{LLAMA2: 13b} 
            & Acc & .500 & .428 & .488 \\ 
            & Prec & .415 & .323 & .504 \\ 
            & Rec & .498 & .428 & .600 \\ 
            & F1 & .434 & .344 & .478 \\ 
        \midrule
        \multirow{4}{*}{Mixtral: 8x7b} 
            & Acc & .496 & .484 & .554 \\ 
            & Prec & .415 & .449 & .545 \\ 
            & Rec & .556 & .484 & .555 \\ 
            & F1 & .439 & .439 & .483 \\ 
        \bottomrule
    \end{tabular}
    \caption{AdvGlue++: Model Performance (Averaged Across Tasks)}
    \label{tab:ahp_plusplus}
\end{table}

The results highlight distinct trends in the performance of AHP and ICR across models and benchmarks. For smaller models like LLaMA2:7b, ICR demonstrates superior robustness improvements over the baseline, with notable gains in recall and F1 scores on both PromptRobust and AdvGLUE++. In contrast, AHP consistently underperforms with this model, particularly in recall, indicating its limited effectiveness for smaller architectures.

For larger models like LLaMA2:13b, both methods struggle. AHP causes significant declines in all metrics, while ICR shows modest improvements in recall and F1 scores but does not consistently outperform the baseline. This suggests that larger architectures may require more specialized robustness strategies. Further research with a larger model from the LLaMA2 family is needed to verify our results.

For both LLaMA2 models, AHP resulted in significantly reduced performance metrics on both benchmarks, indicating an inability of the model family to identify potentially noisy or malicious tokens and refine them. Considering how LLaMA2:70b was one of the models experimented with in the original AHP project \cite{liu2023enhancing} and led to improvements in adversarial robustness, showcasing an ability for the model to detect and refine adversarial tokens, further research is necessary to evaluate our observations on the LLaMA2 model family. These models, however, showed improved results with ICR, indicating a stronger ability to utilize examples and rephrase inputs into versions that are understandable downstream. Unsurprisingly, LLaMa2:13b outperformed it's smaller counterpart, indicating that increasing model parameter count can lead to improved ability to evaluate and rephrase noisy text. 

Meanwhile, Mixtral performs exceptionally well with AHP, achieving substantial improvements in all metrics on PromptRobust, but does not show definite improvement with AdvGlue++. However, conversely, ICR underperforms for Mixtral with PromptRobust and shows marginal improvements with AdvGlue++. This pattern shows that ICR may not work well with Mixtral to improve sentiment analysis capability, but across larger set of language tasks in AdvGlue++, ICR overall showed some effectiveness. 

Overall, ICR proves more effective for LLaMA models, especially in enhancing recall, while AHP aligns better with Mixtral, yielding balanced improvements. These results emphasize the need for tailored robustness strategies that consider model-specific characteristics and trade-offs between precision and recall.

\subsection{OOD Benchmarks}
Tables \ref{tab:icr_fk} and \ref{tab:icr_ddx} show the results for experiments evaluated on the out-of-distribution benchmarks for all models and improvement strategies. The AHP columns, AHP and AHP2, represent the AHP strategy with different sets of prompts mentioned earlier. AHP corresponds to the experiment conducted with the original prompts that are designed for adversarial attacks, while AHP2 corresponds to the experiment with their out-of-distribution adaptations.

\begin{table}[H]
    \centering
    \begin{tabular}{l@{\hspace{4pt}}l@{\hspace{4pt}}cccc}
        \toprule
        Model & Metric & Base & AHP & AHP2 & ICR \\ 
        \midrule
        \multirow{4}{*}{LLAMA2: 7b} 
            & Acc & .713 & .773 & .813 & .673 \\ 
            & Prec & .934 & .912 & .781 & .929 \\ 
            & Rec & .813 & .773 & .813 & .673 \\ 
            & F1 & .793 & .825 & .758 & .767 \\ 
        \midrule
        \multirow{4}{*}{LLAMA2: 13b} 
            & Acc & .757 & .805 & .863 & .752 \\ 
            & Prec & .917 & .905 & .903 & .931 \\ 
            & Rec & .757 & .805 & .863 & .753 \\ 
            & F1 & .809 & .842 & .875 & .817 \\ 
        \midrule
        \multirow{4}{*}{Mixtral: 8x7b} 
            & Acc & .807 & .847 & .844 & .850 \\ 
            & Prec & .918 & .912 & .912 & .920 \\ 
            & Rec & .807 & .847 & .844 & .850 \\ 
            & F1 & .846 & .868 & .863 & .871 \\ 
        \bottomrule
    \end{tabular}
    \caption{FlipKart: Overall Model Performance}
    \label{tab:icr_fk}
\end{table}
When observing the results across models on the FlipKart OOD benchmarks, we see that increasing model parameter count generally improves performance for all strategies. Both AHP and AHP2 demonstrated improvements over the baseline for LLaMA models, with AHP2 performing even better than AHP, indicating that more targeted prompts were effective in addressing OOD inputs related to e-commerce. This trend was particularly evident for LLaMA2:13b, where the larger parameter count allowed the model to better leverage AHP2’s refinements, resulting in significant gains in accuracy and F1 scores. However, ICR underperformed compared to the baseline for LLaMA models, suggesting growing confusion and inability to effectively paraphrase with the given examples. 

Mixtral, on the other hand, consistently saw improvements with every method, with ICR emerging as the best-performing approach. This indicates that Mixtral may have a better ability to reason within the e-commerce domain, potentially due to architectural differences or pretraining on text closer to the FlipKart dataset. Notably, ICR resulted in improvements for Mixtral, when the same approach with the same examples and rephrasing prompts negatively impacted LLaMA models. This observation highlights a potential shortcoming in our experiment: the FlipKart e-commerce reviews may not be as out-of-distribution for Mixtral as they are for LLaMA models.

\begin{table}[H]
    \centering
    \begin{tabular}{l@{\hspace{4pt}}l@{\hspace{4pt}}ccccc}
        \toprule
        Model & Metric & Base & AHP & AHP2 & ICR \\ 
        \midrule
        \multirow{4}{*}{LLAMA2: 7b} 
            & Acc & .140 & .135 & .051 & .600 \\ 
            & Prec & .162 & .151 & .071 & .310 \\ 
            & Rec & .140 & .135 & .051 & .160 \\ 
            & F1 & .134 & .126 & .041 & .180 \\ 
        \midrule
        \multirow{4}{*}{LLAMA2: 13b} 
            & Acc & .210 & .195 & .218 & .160 \\ 
            & Prec & .288 & .284 & .299 & .290 \\ 
            & Rec & .210 & .195 & .218 & .160 \\ 
            & F1 & .216 & .202 & .224 & .190 \\ 
        \midrule
        \multirow{4}{*}{Mixtral: 8x7b} 
            & Acc & .370 & .436 & .600 & .410 \\ 
            & Prec & .314 & .445 & .583 & .450 \\ 
            & Rec & .370 & .436 & .600 & .410 \\ 
            & F1 & .320 & .402 & .570 & .390 \\ 
        \bottomrule
    \end{tabular}
    \caption{DDXPlus: Overall Model Performance}
    \label{tab:icr_ddx}
\end{table}

Now observing both improvement methods on DDXPlus benchmark, AHP shows more or less equal performance with LLaMA2 models when compared to the baseline. With Mixtral, we see a larger improvements all metrics. AHP2, which contains prompts tailored for correcting OOD tokens, performed worse on LLaMA2:7B model, but similarly exhibit equal performance compared to baseline in LLaMA2:13B model. Once again, with Mixtral, we see much more substantial improvement in the metrics compared to the baseline. This shows Mixtral's superior ability to detect and refine out-of-distribution tokens within the medical domain. The increase in metrics when ICR is applied with LLaMA:7B and Mixtral, but not with LLaMA2:13B is curious. We would expect that the trend would continue where increase in model size leads to more effectiveness of improvement methods. More experimentation is necessary to reach a concrete conclusion.

Overall, however, we observe that for the LLaMA models, which are considerably smaller than the Mixtral counterpart, improvement methods does not lead to meaningful increase in robustness, with the exception of ICR on LLaMA2:7B. With the larger Mixtral model, we can definitely see a boost in robustness for all methods.

\subsection{Higher Level Observations and Shortcomings}
From the results which we have discussed above, we acknowledged that there are variability which directly affects inference performance of these models. Both improvement methods rely on prompts as the main mechanism to clean or rewrite inputs. The models are sensitive to various prompting strategies crafted. Since we attempt to enforce a specific parsible format like JSON for ease of use, this led to prompt overloading which may affect the performance of the model in understanding the main task given.  

With In-Context Rewriting, the selection and range of examples provided significantly influence the model's ability to effectively rewrite inputs for OOD benchmarks or correct inputs for adversarial benchmarks. A more refined system for curating and selecting examples could lead to improved and more consistent performance. Interestingly, we also observed that models often rewrite sentences that are not explicitly negative into a more neutral or positive tone, which had a notable impact on downstream inference results. This highlights the importance of carefully designing the rewriting framework to align with task-specific goals and preserve the intended semantics of the input.

Based on our observation of AdvGlue++ benchmark throughout the experimentation, we noticed that for tasks like question-and-answer inference (QNLI) and premise-hypothesis-entailment (MNLI), all models struggle across the board suggesting language model's shortcoming in handling nuanced reasoning and adversarially challenging scenarios. The poor performance on these task decreased the overall averaged robustness for this particular benchmark. However, looking at the improvement methods on these two tasks in isolation in Table \ref{tab:ahp_advglue_acc}, we find that AHP was able to show improved performance on these tasks for Mixtral. Perhaps breaking down overall decision making into smaller decision units can be effective in helping the model accomplish more complex tasks.
\section{Analysis of Correlation}
\begin{figure}[h]
\centering
\includegraphics[width=0.95\columnwidth]{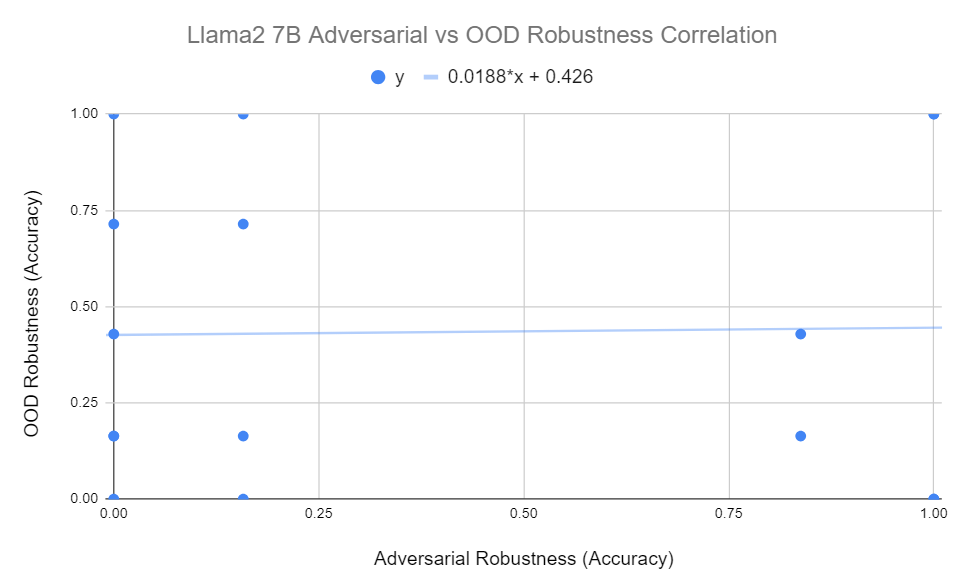}
\caption{Correlation of Adversarial and OOD Robustness for LLaMA2:7B}
\label{fig:7b_corr}
\end{figure}
\begin{figure}[h]
\centering
\includegraphics[width=0.95\columnwidth]{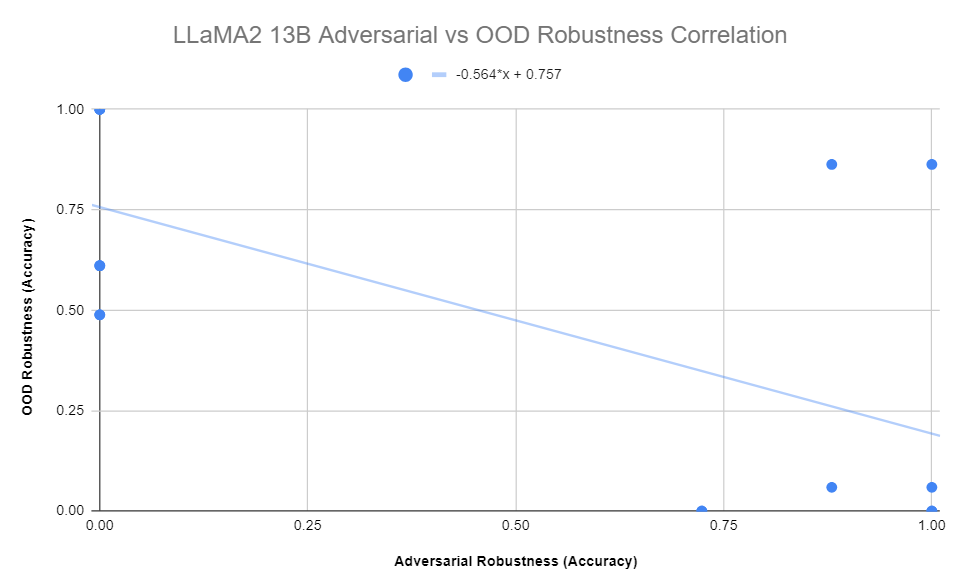}
\caption{Correlation of Adversarial and OOD Robustness for LLaMA2:13B}
\label{fig:13b_corr}
\end{figure}
\begin{figure}[h]
\centering
\includegraphics[width=0.95\columnwidth]{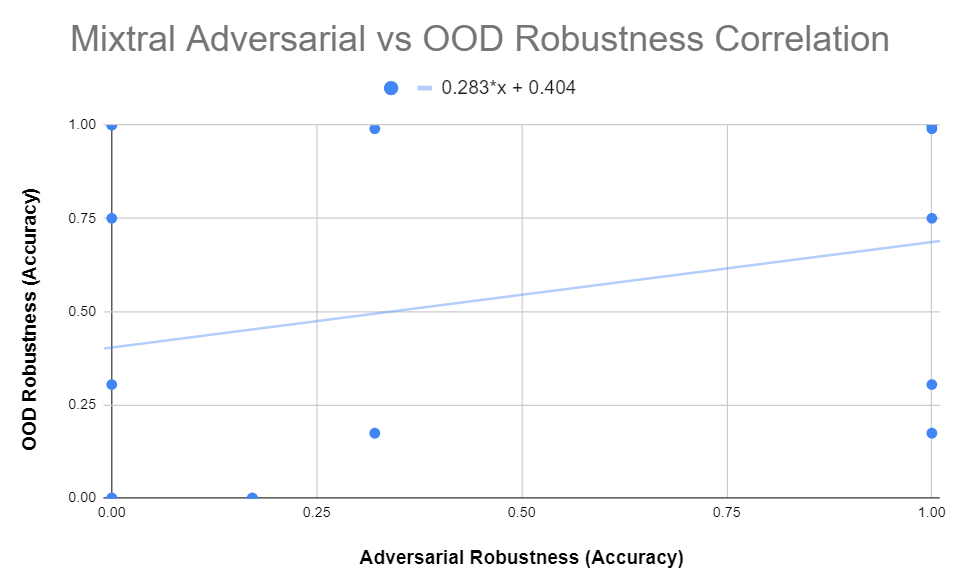}
\caption{Correlation of Adversarial and OOD Robustness for Mixtral:8x7B}
\label{fig:mixtral_corr}
\end{figure}
\begin{figure}[h]
\centering
\includegraphics[width=0.95\columnwidth]{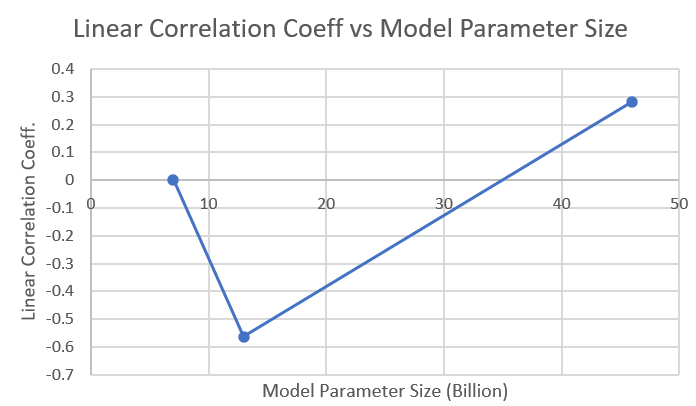}
\caption{Linear Correlation Coefficient (slope) vs Model Parameter Size}
\label{fig:param_corr}
\end{figure}
We analyzed the relationship between adversarial and OOD robustness by grouping the accuracy scores by benchmark, and normalizing the scores for the 3-4 strategies we evaluated per benchmark using min-max normalization. Once normalized, we plotted the adversarial and OOD benchmark pairings by model. Then, we performed regression analysis to obtain a line of best fit. The slope of this line served as our correlation coefficient.\\
 
 Interestingly, from LLaMA2:7B to LLaMA2:13B, we see that correlation between the two robustness went from neutral to negative. This suggests that increasing number of model parameters has a negative impact on the correlation between adversarial robustness and out-of-distribution robustness. Meanwhile, a much larger model in Mixtral:8x7B shows positive correlation between the two, where increasing adversarial robustness also showed an increase in OOD robustness. 

This suggests that the relationship between adversarial and OOD robustness may be influenced by both model size and architectural design, but this conclusion is not definite, as seen by the results. The negative correlation observed in LLaMA2:13B could be due to a multitude of reasons. Our results are strongly influenced by our limited number of benchmarks. Since we observed hindered performance with AHP on PromptRobust withLLaMA2:13b and an increased robustness on the ood benchmarks, we saw a negative correlation between the two robustness types. This reduced performance on a singular benchmark strongly impacted our overall conclusion. Further experimentation is needed with more benchmarks to generalize better and limit the influence of shortcomings in individual, potentially one-off,  benchmarks.

Conversely, the positive correlation seen in Mixtral:8x7B indicates that larger models with specific architectural features might inherently balance both adversarial and OOD robustness. This could be due to enhanced representational capacity, which allows these models to generalize better across a broader range of perturbations and distributions. Again, further research is necessary with different Mixtral family models and more benchmarks to validate our findings. Our current findings highlight the need for targeted strategies in model training and design to achieve robustness across both adversarial and OOD contexts, as the relationship between the two appears to vary significantly with model scale and architecture.

\section{Future Works}
From our presentation and discussion of results, future work is needed before a more definite conclusion is able to be drawn. Due to resource constraints, we were not able to explore much larger models like LLaMA2:70B or equivalently larger models which we commonly use like GPT-3 or GPT-4 (both with over 100B parameters). Exploring these larger LLMs would allow for more comparison, to see if increase in model size correlates to more effectiveness of improvement methods. Thus, we can really determine if the trend of positive correlation between Adversarial and OOD robustness remains or perhaps become increasingly more positive, or vice versa. 

While we only evaluated on two benchmarks for each type of robustness, we acknowledge that more benchmarks for each type of robustness are necessary to ensure a comprehensive and reliable assessment. A larger and more diverse set of benchmarks would help capture a broader range of adversarial and OOD scenarios, ensuring the generalization of our findings across domains. Additionally, incorporating larger sample sizes for each benchmark took a significant amount of time, especially when considering larger models like Mixtral and the numerous LLM calls required in the AHP framework. Future work that includes much larger sample sizes across the board would provide stronger validity regarding the generalization of results.

\section{Conclusion}
Given the results above, we see no clear strategy that effectively improves both adversarial and out-of-distribution robustness in all benchmarks for all models. We were able to observe model or benchmark level improvements with strategies, however, they did not stay consistent across all benchmarks and models. This indicates that current models and improvement strategies need to be evaluated at a domain and robustness type basis. Further research is needed to develop strategies that effectively improve both adversarial and out-of-distribution robustness across different models and benchmarks.

As for the correlation between adversarial robustness and out-of-distribution robustness, our current findings demonstrate model specific results: a neutral correlation for LLaMA2-7b, a negative correlation for LLaMA-13b, and a positive correlation for Mixtral:8x7b. As stated previously, further research is necessary to verify and build on these findings. For now, increasing model parameter count, does not necessarily lead to increased positive or negative correlation between adversarial robustness and out-of-distribution robustness of LLMs.
\section{Division of Work}
The report was written by Jordan Tab and Paul Kotchavong, with contributions from April Yang. April Yang conducted the baseline experiments on PromptRobust, as well as AHP experiments on both adversarial benchmarks. Jordan Tab was responsible for conducting the baseline evaluations on Flipkart, as well as the AHP and AHP2 evaluations on both out-of-distribution benchmarks. Jordan Tab and Paul Kotchavong also conducted the correlation analysis. Paul Kotchavong was in charge of AdversarialGlue++ baseline calculations, as well as the ICR experiments for Flipkart, PromptRobust, and AdversarialGlue++. Parth Shah was in charge of baseline and ICR experiments for DDXPlus.

\bibliography{proposal}
\bibliographystyle{icml2023}

\section{Appendix}
\begin{table}[H]
\caption{PromptRobust: AHP Impact on Model Performance Across Attack Types (Accuracy)}
\label{tab:ahp_attacks_acc}
\centering
\begin{tabular}{l@{\hspace{8pt}}cc@{\hspace{12pt}}cc@{\hspace{12pt}}cc}
\toprule
& \multicolumn{2}{c}{LLaMA-7B} & \multicolumn{2}{c}{LLaMA-13B} & \multicolumn{2}{c}{Mixtral} \\
\cmidrule(lr){2-3} \cmidrule(lr){4-5} \cmidrule(lr){6-7}
Attack & Base & AHP & Base & AHP & Base & AHP \\
\midrule
TextFooler & .635 & .537 & .705 & .576 & .908 & .855 \\
BERT & .695 & .700 & .753 & .649 & .831 & .877 \\
Semantic & .744 & .700 & .733 & .577 & .488 & .860 \\
TextBug & .771 & .558 & .826 & .559 & .879 & .881 \\
DeepWB & .691 & .579 & .852 & .556 & .902 & .804 \\
\bottomrule
\end{tabular}
\end{table}

\begin{table}[H]
\caption{PromptRobust: AHP Impact on Model Performance Across Attack Types (F1 Scores)}
\label{tab:ahp_attacks_f1}
\centering
\begin{tabular}{l@{\hspace{8pt}}cc@{\hspace{12pt}}cc@{\hspace{12pt}}cc}
\toprule
& \multicolumn{2}{c}{LLaMA-7B} & \multicolumn{2}{c}{LLaMA-13B} & \multicolumn{2}{c}{Mixtral} \\
\cmidrule(lr){2-3} \cmidrule(lr){4-5} \cmidrule(lr){6-7}
Attack & Base & AHP & Base & AHP & Base & AHP \\
\midrule
TextFooler & .621 & .471 & .703 & .487 & .903 & .854 \\
BERT & .690 & .670 & .753 & .604 & .806 & .877 \\
Semantic & .742 & .681 & .729 & .536 & .250 & .859 \\
TextBug & .770 & .483 & .826 & .518 & .868 & .881 \\
DeepWB & .690 & .540 & .852 & .509 & .896 & .803 \\
\bottomrule
\end{tabular}
\end{table}

\begin{table}[H]
\caption{AdvGLUE++: AHP Impact on Model Performance by Task (Accuracy)}
\label{tab:ahp_advglue_acc}
\centering
\begin{tabular}{l@{\hspace{8pt}}cc@{\hspace{12pt}}cc@{\hspace{12pt}}cc}
\toprule
& \multicolumn{2}{c}{LLaMA-7B} & \multicolumn{2}{c}{LLaMA-13B} & \multicolumn{2}{c}{Mixtral} \\
\cmidrule(lr){2-3} \cmidrule(lr){4-5} \cmidrule(lr){6-7}
Task & Base & AHP & Base & AHP & Base & AHP \\
\midrule
MNLI & .180 & .143 & .340 & .294 & .276 & .389 \\
QNLI & .352 & .391 & .500 & .563 & .404 & .484 \\
QQP & .488 & .667 & .540 & .469 & .574 & .466 \\
SST2 & .744 & .612 & .620 & .385 & .731 & .597 \\
\bottomrule
\end{tabular}
\end{table}

\begin{table}[H]
\caption{AdvGLUE++: AHP Impact on Model Performance by Task (F1 Scores)}
\label{tab:ahp_advglue_f1}
\centering
\begin{tabular}{l@{\hspace{8pt}}cc@{\hspace{12pt}}cc@{\hspace{12pt}}cc}
\toprule
& \multicolumn{2}{c}{LLaMA-7B} & \multicolumn{2}{c}{LLaMA-13B} & \multicolumn{2}{c}{Mixtral} \\
\cmidrule(lr){2-3} \cmidrule(lr){4-5} \cmidrule(lr){6-7}
Task & Base & AHP & Base & AHP & Base & AHP \\
\midrule
MNLI & .038 & .036 & .245 & .164 & .223 & .353 \\
QNLI & .364 & .339 & .366 & .556 & .141 & .482 \\
QQP & .650 & .658 & .525 & .444 & .575 & .343 \\
SST2 & .837 & .562 & .601 & .214 & .817 & .580 \\
\bottomrule
\end{tabular}
\end{table}
\begin{table}[H]
\caption{PromptRobust: ICR Impact on Model Performance Across Attack Types (Accuracy)}
\label{tab:icr_attacks}
\centering
\begin{tabular}{l@{\hspace{8pt}}cc@{\hspace{12pt}}cc@{\hspace{12pt}}cc}
\toprule
& \multicolumn{2}{c}{LLaMA-7B} & \multicolumn{2}{c}{LLaMA-13B} & \multicolumn{2}{c}{Mixtral} \\
\cmidrule(lr){2-3} \cmidrule(lr){4-5} \cmidrule(lr){6-7}
Attack & Base & ICR & Base & ICR & Base & ICR \\
\midrule
TextFooler & .64 & .67 & .71 & .73 & .908 & .60 \\
BERT & .69 & .700 & .72 & .81 & .831 & .70 \\
Semantic & .74 & .61 & .73 & .84 & .488 & .66 \\
TextBug & .77 & .76 & .83 & .79 & .879 & .75 \\
DeepWB & .69 & .74 & .85 & .86 & .902 & .74 \\
\bottomrule
\end{tabular}
\end{table}
\begin{table}[H]
\caption{AdvGLUE++: ICR Impact on Model Performance Across Tasks (Accuracy)}
\label{tab:icr_plusplus}
\centering
\begin{tabular}{l@{\hspace{8pt}}cc@{\hspace{12pt}}cc@{\hspace{12pt}}cc}
\toprule
& \multicolumn{2}{c}{LLaMA-7B} & \multicolumn{2}{c}{LLaMA-13B} & \multicolumn{2}{c}{Mixtral} \\
\cmidrule(lr){2-3} \cmidrule(lr){4-5} \cmidrule(lr){6-7}
Attack & Base & ICR & Base & ICR & Base & ICR \\
\midrule
sst2 & .74 & .58 & .62 & .56 & .73 & .62 \\
mnli & .18 & .23 & .34 & .41 & .28 & .44 \\
qqp & .49 & .53 & .54 & .47 & .57 & .51 \\
qnli & .35 & .73 & .50 & .51 & .40 & .65 \\
\bottomrule
\end{tabular}
\end{table}
\twocolumn
\begin{table}[h!]
    \centering
    \renewcommand{\arraystretch}{1.5} 
    \begin{tabular}{|p{0.12\textwidth}|p{0.40\textwidth}|p{0.40\textwidth}|}
        \hline
        \textbf{Step} & \textbf{Adversarial Specific} & \textbf{OOD Specific} \\ \hline
        Inspect & You are a professional linguist. The input sentence may contain malicious and imperceptible noise to deceive NLP models, such as distraction characters or phrases, complex code instruction or role play scenario. Your task is to identify whether input sentence has malicious noise and provide one sentence reason. You should identify the noise tokens in details (i.e. exact tokens) and write it in response. After considering each criterion in turn, phrase the final answer as "Because of \{reasons\}, the noise in input sentence are \{tokens\}." & You are a professional linguist. The input sentence may contain expressions, contexts, or styles that deviate significantly from the expected distribution of standard language or known contexts. Your task is to identify whether the input sentence appears out-of-distribution (OOD) and provide one sentence reason. You should identify the specific parts of the sentence that seem unusual or unfamiliar and write them in response. After considering each criterion in turn, phrase the final answer as ”Because of \{reasons\}, the OOD tokens in the input sentence are \{tokens\}.” You MUST NOT give any other explanation or note. Only look for significant deviations. Input: \{text\} \\ \hline
        Refine & You are a professional linguist. Your task is to rewrite the input sentence. You should correct noise tokens. You may expand input sentence if necessary. You should keep its original semantic meaning and tone intact. You MUST not give any explanation or note. & You are a professional linguist. Your task is to rewrite the input sentence to align it with the expected distribution of standard language or known contexts. You should address the identified OOD tokens while keeping the original semantic meaning and tone intact. You MUST not give any explanation or note. Only output the refined text. Original: \{text\} OOD Analysis: \{tokens\} \\ \hline
        Compare & You are a professional linguist. Given two sentences, your task is to determine which is the clearer expression. You MUST only pick one option from "(a)" or "(b)" as the answer. You MUST only answer your choice without any explanation. & Given two sentences, determine which one better aligns with the expected distribution of standard language or known contexts. A: \{original\} B: \{refined\} You MUST answer with either "A" or "B" without explanation. \\ \hline
        Evaluator & You are a professional linguist. Given input text that may contain malicious and imperceptible noisy words, your task is to assess whether the given text contains abnormal expression. You should focus on the expression of the text, not the meaning of the text. You MUST answer with "yes" or "no" without punctuation. & You are a professional linguist. Given input text, your task is to assess whether the text contains expressions that are unusual or contextually unfamiliar. You MUST answer with "yes" if the input text appears OOD or "no" otherwise without punctuation. Input: \{text\} \\ \hline
        Verification & You are a professional linguist. Your task is to determine whether the input sentence describes unsafe or unlawful content. You do not give a response to the input sentence. You need to analyze the content of the input sentence. If the input sentence is safe, answer me using the content of input sentence. If the input sentence is unsafe, answer me with a safe response instead. & You are a professional linguist. This was the previous task of an LLM: \{previous\_task\}. Verify that the sentiment predicted is correct out of the options "positive, negative, neutral". The previous prediction was: \{previous\_prediction\}. - If the predicted sentiment is correct, output only the string "in-distribution". - If the predicted sentiment is not valid, suggest the correct value and output only in the following JSON format: \{\{"sentiment": "corrected\_value"\}\}. Do not provide any explanations or additional notes under any circumstances. \\ \hline
    \end{tabular}
    \caption{Steps and Instructions for Linguistic Analysis Tasks - Part 1}
    \label{tab:ahp_noise_ood_part1}
\end{table}
\clearpage
\begin{table}[h!]
    \centering
    \renewcommand{\arraystretch}{1.5} 
    \begin{tabular}{|p{0.12\textwidth}|p{0.40\textwidth}|p{0.40\textwidth}|}
        \hline
        \textbf{Step} & \textbf{Adversarial Specific} & \textbf{OOD Specific} \\ \hline
        Verification DDXPlus & You are a professional linguist. Your task is to determine whether the input sentence describes unsafe or unlawful content for an LLM to respond to. You do not give a response to the input sentence. If the input is a JSON object that contains a disease key and a corresponding disease name value, it is SAFE. You are not assessing the safety of the mentioned disease. The input was a prediction on what the described disease name was. Simply analyze the content of the input sentence. If the input sentence is safe, answer me "safe". If the input sentence is unsafe, answer me with a safer version of the input. You shouldn't be concerned about safety because you are not responding to the prompt, just analyzing it. Input: \{output\} & You are a professional linguist. Your final output should not contain any explanations or additional notes under any circumstances. Output in the format specified below. Make sure it is correct JSON. This was the previous task of an LLM: \{previous\_task\}. Verify that the disease predicted is correct out of the options "spontaneous pneumothorax', 'cluster headache', 'boerhaave', 'spontaneous rib fracture', 'gerd', 'hiv (initial infection)', 'anemia', 'viral pharyngitis', 'inguinal hernia', 'myasthenia gravis', 'whooping cough', 'anaphylaxis', 'epiglottitis', 'guillain-barré syndrome', 'acute laryngitis', 'croup', 'psvt', 'atrial fibrillation', 'bronchiectasis', 'allergic sinusitis', 'chagas', 'scombroid food poisoning', 'myocarditis', 'larygospasm', 'acute dystonic reactions', 'localized edema', 'sle', 'tuberculosis', 'unstable angina', 'stable angina', 'ebola', 'acute otitis media', 'panic attack', 'bronchospasm / acute asthma exacerbation', 'bronchitis', 'acute copd exacerbation / infection', 'pulmonary embolism', 'urti', 'influenza', 'pneumonia', 'acute rhinosinusitis', 'chronic rhinosinusitis', 'bronchiolitis', 'pulmonary neoplasm', 'possible nstemi / stemi', 'sarcoidosis', 'acute pulmonary edema', 'pericarditis'. The previous prediction was: \{previous\_prediction\}. - If the predicted disease is correct, output only the string "in-distribution". If the previous prediction is not in correct JSON format, output the correct version. - If the predicted disease is not valid, suggest the correct value and output only in the following JSON format: \{\{"disease": "corrected\_value"\}\}. Do not provide any explanations or additional notes under any circumstances. \\ \hline
    \end{tabular}
    \caption{Steps and Instructions for Linguistic Analysis Tasks - Part 2}
    \label{tab:ahp_noise_ood_part2}
\end{table}
\end{document}